\documentclass{article}

\usepackage{arxiv}

\usepackage[utf8]{inputenc} 
\usepackage[T1]{fontenc}    
\usepackage{hyperref}       
\usepackage{url}            
\usepackage{booktabs}       
\usepackage{amsfonts}       
\usepackage{nicefrac}       
\usepackage{microtype}      
\usepackage{lipsum}         
\usepackage{graphicx}
\usepackage{doi}
\usepackage{xcolor}
\usepackage{makecell}
\usepackage{amsmath}
\usepackage{cleveref}
\usepackage[
  backend=biber,
  style=authoryear,
  natbib=true,
  maxbibnames=10,
  minbibnames=10,
  maxcitenames=1,
  mincitenames=1,
  uniquename=false,
  uniquelist=false
]{biblatex}

\AtEveryCite{%

}

\title{Made in Hungary: Comments on the performance of generative language models}

\renewcommand{\shorttitle}{Comments on Hungarian Language Models}

\author{%
\normalfont\normalsize
\begin{tabular}{c}
  \textbf{Mátyás~Osváth}\hspace{2.5em}%
  \textbf{Máté Molnár}\hspace{2.5em}%
  \textbf{Enikő~Héja}\hspace{2.5em}%
  \textbf{Noémi~Ligeti-Nagy}\\[2ex]
  Institute for Language Technologies and Applied Linguistics\\
  ELTE Research Centre for Linguistics\\
  Budapest, Hungary\\[1ex]
  \texttt{\{osvath.matyas,\,molnar.mate,\,heja.eniko,\,ligeti-nagy.noemi\}@nytud.elte.hu}
\end{tabular}%
}

\hypersetup{
pdftitle={Made in Hungary: Comments on the performance of generative language models},
pdfsubject={cs.CL},
pdfauthor={Mátyás Osváth, Máté Molnár, Enikő Héja, Noémi Ligeti-Nagy},
pdfkeywords={hungarian language models, generative language models, low-resource languages},
}

\begin{document}
\maketitle

\begin{abstract}
In recent years, three initiatives have emerged to develop generative language models in Hungary.
The motivation behind them is the same.
For Hungarian, no model with the given capability existed, or existing English-centric models offered limited proficiency.
A detailed examination of the corresponding studies, however, reveals several methodological limitations.
First, the reliability of the evaluation protocols is questionable.
Contrary to the findings of Csibi et al. [2026], evaluation under the recommended inference settings shows that \textsc{Qwen3-4B} achieves higher scores than \textsc{Racka-4B}, its Hungarian-adapted version.
Data contamination is evident in the work of Yang et al. [2025d] and Szentmihályi et al. [2025], potentially biasing the reported results.
Second, the training pipelines fall short of current best practices in corpus curation and data mixture, which risks wasting substantial compute on low-quality data.
The lack of controlled ablations prevents reliable assessment of these choices.
Third, none of the three papers assessed forgetting or capability loss.
Testing the adapted models on a subset of the original benchmarks indicates performance decline in all three cases, especially \textsc{Racka-4B}.
These observations emphasize the importance of rigorous experimental design in language model development, given the significant computational and financial costs involved.
\end{abstract}


\section{Introduction}

Over the past decade, progress in artificial intelligence -- and in language technology in particular -- has accelerated drastically,
and has passed through several stages of development (see Appendix~\ref{appendix:stateofllms} for a brief overview).
Three Hungarian initiatives have emerged in recent years to build generative language models for Hungarian, each following a different methodological approach.
In addition, several international initiatives have produced models adapted to Hungarian or with Hungarian capability \citep{salamandra_2025, csaki_sambalingo_2024, salamanca_tiny_2026}, but these fall outside the scope of the present analysis.
The first approach, taken by \citet{otp_2025}, follows the architecture of \textsc{GPT-3} \citep{brown2020languagemodelsfewshotlearners} to build bilingual pretrained models in an industrial collaboration, which they named \textsc{OTP-1.5B} and \textsc{OTP-13B}.
The second approach combined continual pretraining with supervised fine-tuning. The models were adapted from \textsc{Llama 3.1} \citep{grattafiori2024llama3herdmodels} and \textsc{Qwen2.5} \citep{qwen2025qwen25technicalreport} chat models. The authors first utilized  continual pretraining on Hungarian and multilingual corpora and then fine-tuned the models on instruction-following and chat data\footnote{The ELTE Hungarian Research Centre for Linguistics has previously released several Hungarian generative models, but this analysis focuses on the two most recent chat models.} \citep{chatpuli_2025, yang2025puli}.
The third is the \textsc{Racka-4B} model \citep{racka_2026}, introduced in 2026, where parameter-efficient training was applied to the \textsc{Qwen3-4B} reasoning model \citep{yang2025qwen3technicalreport}.
Specifically, continued pretraining was performed with low-rank adaptation (LoRA) \citep{hu2021lora}, and the authors also adapted the tokenizer to Hungarian.
The three approaches therefore represent three distinct points on the language-adaptation spectrum.

The motivations behind these works are largely the same. All three works start from the premise that the Hungarian performance of English-centric models is limited by the underrepresentation of the language.
A further motivation -- stated explicitly by \textsc{ChatPULI} and \textsc{Racka} papers -- is that no model with the given capability, chat and reasoning respectively, was available for Hungarian at the time.

All three initiatives remain within the earlier stages of the timeline, namely pretraining and supervised fine-tuning.
If the comparison is extended to models developed by international teams, it becomes clear that the focus of Hungarian language modelling has shifted.
International models already outperformed the Hungarian ones on Hungarian tasks, even at the time of their release.

We argue that systematic shortcomings can be identified in the development of these models, from data preparation to evaluation.
To support this claim, we show that the evaluation protocols are inadequate and that reevaluating the models changes the outcome (Section~\ref{sec:evaluation}).
In addition, corpus curation and data mixture design do not reflect current best practices and thus likely contribute to these results (Sections~\ref{sec:training} and~\ref{sec:ablations}).
We also assess forgetting and capability loss, which none of the studies examined (Section~\ref{sec:forgetting}).
Finally, the training cost and carbon footprint of the models are estimated in order to put the identified issues into perspective (Section~\ref{sec:trainingcost}).

\section{The reliability of reported results}
\label{sec:evaluation}

In the three study examined, assessing model capabilities through benchmarks faces challenges on several fronts: contamination of the train-test separation, evaluation protocols that render results incomparable across papers, and benchmarks whose validity for measuring Hungarian capability is itself questionable.

\textsc{ChatPULI} explicitly reports incorporating $7,344$ \textsc{HuLU} prompts into its supervised fine-tuning data\footnote{The paper does not specify whether these prompts were used in modified form or not.} and attributes part of the models' improvement to prior exposure to the benchmark -- a case of instance-level contamination, which inflates scores even when the overlap involves rephrased rather than verbatim test items \citep{palavalli2024taxonomydatacontaminationlarge}. The other two publications neither report a decontamination procedure nor discuss train-test overlap, leaving the extent of contamination unknown in their case.

\textsc{Racka}'s performance on the \textsc{HuLU} benchmark \citep{ligetinagy2024hulu} is measured on the public validation split rather than the held-out test set used by the official leaderboard\footnote{\url{https://hulu.nytud.hu/leaderboard}} -- a choice stated only in the results table caption -- which renders the reported figures incomparable to published \textsc{HuLU} results. They also state that they modified the evaluation scripts\footnote{\url{https://github.com/nytud/HuLU/tree/main/evaluate}}, but do not provide details, which makes it unlikely to reproduce their results\footnote{It is possible that the CoPA task was modified, since the \texttt{AutoModelForMultipleChoice} class used in the evaluation is only implemented for encoder models in the \texttt{transformers} library.}.

The authors of the \textsc{OTP} models also show evidence of data contamination in their evaluation. Two of the three hand-crafted benchmarks were constructed directly from the authors' own pretraining corpus. Specifically, forum comments for the sentiment task and tagged news articles for the topic classification task. The evaluation items are therefore drawn from the same data distribution the models were trained on, while competing systems could be trained on different corpora, resulting in an unfair advantage over other models. The paper does not report whether the specific documents used for evaluation were excluded from training. A further observation is that every evaluated model scores exactly $27.5$ on the integrity task, indicating that the task does not discriminate between systems.

In the \textsc{Racka} study, the \textsc{OpenHuEval} \citep{openhueval} evaluation was conducted with reasoning disabled -- a configuration disclosed only in the appendix -- and applied -- as far as can be determined from the description -- to both \textsc{Racka} and the \textsc{Qwen3-4B} model. The stated motivation is the incompatibility of the benchmark implementation with reasoning output and the tendency of small reasoning models to enter repetition loops under greedy decoding (with temperature $0.0$)\footnote{\url{https://qwen.readthedocs.io/en/v3.0/getting_started/quickstart.html}}. Both of them are addressable, the former by stripping reasoning traces before scoring, the latter by using the model's recommended sampling parameters and averaging over multiple runs.
The \textsc{Qwen3-4B} model card (and paper) explicitly recommends generation parameters for reasoning mode and warns against greedy decoding\footnote{\url{https://huggingface.co/Qwen/Qwen3-4B\#best-practices}}.

The team behind \textsc{Racka} excludes international models from its comparison on the grounds that these ``according to our experiences tend to perform lower than models trained in Hungary'' \citep[Sec.~5]{racka_2026}, without supporting measurement, also noting that a broader comparison is left to future work. The claim is difficult to reconcile with the results of \textsc{OpenHuEval}, the benchmark the paper itself adopts, in which \textsc{DeepSeek-R1} \citep{deepseek_r1_2025} and \textsc{GPT-4o} \citep{openai2024gpt4ocard} lead across the Hungarian tasks. The reported figures are also more mixed than the aggregate suggests. Of the eight \textsc{OpenHuEval} subtasks, \textsc{Racka} achieves the best result on only one and matches \textsc{PULI-LlumiX-Llama-3.1} on a second, while the \textsc{Qwen3-4B} baselines lead on three and \textsc{PULI} on three tasks. A similar claim appears in \textsc{ChatPULI} study, where the authors report that continual pretraining led the model to generate ``more grammaticality errors'' and introduce a corrective Hungarian-only training stage in response, without reporting an error rate before or after, or specifying how the degradation was assessed. In both cases, a claim that could have been measured is asserted instead.

Notably, the Hungarian versions of the \textsc{ARC}, \textsc{MMLU}, \textsc{HellaSwag}, \textsc{TruthfulQA} and \textsc{GSM8K} benchmarks used in the final part of the \textsc{Racka} evaluation were produced by machine translation alone, without manual validation \citep{lai2023okapi, thellmann2024multilingualllmevaluationeuropean}.
However, a more significant methodological issue is that the first four benchmarks were evaluated using the default parameter settings of \texttt{lm-eval-harness} \citep[Sec.~4.5]{racka_2026}.
According to the publicly available task configurations\footnote{\url{https://github.com/EleutherAI/lm-evaluation-harness/tree/main/lm_eval/tasks/okapi}}, all four benchmarks use the \texttt{multiple\_choice} output type, which ranks the answer options by log-likelihood\footnote{\url{https://lm-evaluation-harness.readthedocs.io/api/task_configuration/}}.
This means that models -- especially reasoning models -- cannot generate explicit, autoregressive reasoning traces before outputting their final answer.\footnote{These settings can be overridden manually in \texttt{lm-eval-harness}. We used the \texttt{generate\_until} output type in our own evaluation, under which the model's reported performance could be reproduced.}
Furthermore, it can be assumed that the \texttt{'Please reason step by step, and put your final answer within \textbackslash boxed\{\}.'}instruction recommended for evaluating \textsc{Qwen3-4B} was not included in the prompt for the \textsc{GSM8K} benchmark, since it is not part of the chat template and the study does not mention adding it.
The results appear to be consistent with this methodological concern. In several cases, there is no substantial difference between \textsc{Qwen3-4B-Base}, \textsc{Qwen3-4B} and \textsc{Racka-4B}, and \textsc{Qwen3-4B-Base} even outperforms \textsc{Qwen3-4B} on two metrics, achieving the best average performance overall.
The comparison against \textsc{PULI} is also made with an older model, called \textsc{PULI-Llumix-Llama-3.1-8B}, which is not the base model of \citet{chatpuli_2025}, and not its chat version either.
Consequently, the reported results are difficult to interpret, and we argue that the measurements should be repeated with the settings described above.


Human evaluation is likewise absent throughout the studies. Elsewhere this gap is filled by arena-style platforms, as used in \citet{grattafiori2024llama3herdmodels, thellmann2024multilingualllmevaluationeuropean}. We are not aware of such a platform for Hungarian, and \textsc{ChatPULI}'s conversational capability rests on hand-selected dialogues \citep[Table~6]{chatpuli_2025}.

Although the \textsc{Racka} paper presents the model as the first Hungarian reasoning model, they do not provide thorough evaluation of its reasoning capabilities. In particular, the authors do not assess the added value of reasoning for Hungarian tasks or examine whether these capabilities are preserved following continual pretraining.

To address these inconsistencies in the evaluation methodology, we reevaluate all base and adapted models on the translated \textsc{ARC}, \textsc{MMLU}, \textsc{HellaSwag}, \textsc{TruthfulQA} and \textsc{GSM8K} benchmarks (Table \ref{tab:results}), as well as on the \textsc{OpenHuEval} suite (Table \ref{tab:results_openhueval}).
In addition, we also assess the models on the \textsc{HuGME} benchmark \citep{ligetinagy2025hugme}, which is a more recent and comprehensive evaluation suite for generative language models (see Table \ref{tab:hugme_racka} and Table \ref{tab:hugme_puli}).
We do not include \textsc{HuLU} in this comparison, as it was originally designed for encoder-based models.
The \textsc{OTP} models are excluded because the model weights have not been released and no public inference endpoint is available, preventing independent evaluation.
The parameters for the evaluations can be found in Appendix \ref{app:sampling_params_hugme}.

\begin{table*}[ht]
\centering
\setlength{\tabcolsep}{7pt}
\begin{tabular}{@{}l cc cc cc@{}}
\toprule
\textbf{Benchmark}
& \makecell[c]{Qwen3-4B} & \makecell[c]{Racka-4B} & \makecell[c]{Llama 3.1 8B} & \makecell[c]{ChatPULI} & \makecell[c]{Qwen2.5 7B} & \makecell[c]{PULI \\ Trio-Q Chat} \\
\midrule
ARC            & \textbf{88.12\tiny$\pm$0.47} & 74.33\tiny$\pm$1.51 & 69.30\tiny$\pm$0.68 & 66.51\tiny$\pm$0.49 & 58.90\tiny$\pm$0.48 & 73.68\tiny$\pm$0.88 \\
MMLU           & \textbf{71.53\tiny$\pm$0.23} & 54.76\tiny$\pm$0.31 & 52.44\tiny$\pm$0.14 & 50.02\tiny$\pm$0.21 & 48.74\tiny$\pm$0.11 & 54.08\tiny$\pm$0.15 \\
HellaSwag      & \textbf{57.40\tiny$\pm$0.36} & 40.99\tiny$\pm$0.51 & 43.68\tiny$\pm$0.35 & 44.61\tiny$\pm$0.18 & 46.56\tiny$\pm$0.11 & 50.50\tiny$\pm$0.61 \\
TruthfulQA {\scriptsize (MC1)}  & \textbf{56.52\tiny$\pm$1.00} & 41.60\tiny$\pm$1.20 & 53.51\tiny$\pm$0.76 & 44.98\tiny$\pm$1.23 & 51.80\tiny$\pm$0.35 & 52.71\tiny$\pm$1.20 \\
GSM8K {\scriptsize (0-shot, CoT)}  & \textbf{81.56\tiny$\pm$0.20} & 64.66\tiny$\pm$1.43 & 55.00\tiny$\pm$1.13 & 41.70\tiny$\pm$1.37 & 58.04\tiny$\pm$1.24 & 56.48\tiny$\pm$0.80 \\
\bottomrule
\end{tabular}
\caption{Performance of base models and their Hungarian-adapted counterparts on the translated benchmarks. Reasoning models are set to thinking mode. Mean $\pm$ standard deviation over 5 runs. Best result per row in bold.}
\label{tab:results}
\end{table*}

The results in Table~\ref{tab:results} differ substantially from those reported in the \textsc{Racka} paper \citep[Table~3]{racka_2026}.
While the reported average scores of \textsc{Qwen3-4B} and \textsc{Racka-4B} are practically indistinguishable (45.3 and 45.4, respectively), the corrected evaluation protocol reveals an average gap of 15.76 percentage points in favor of \textsc{Qwen3-4B}.
\textsc{ChatPULI} not only falls behind \textsc{Qwen3-4B}, a model half its size,
but also underperforms relative to its own base model, \textsc{Llama 3.1 8B}, on four of the five benchmarks.
In contrast, \textsc{PULI Trio-Q Chat} is the only adapted model that improves over its base model, outperforming it on four of the five benchmarks.

\begin{table*}[ht]
\centering
\setlength{\tabcolsep}{7pt}
\begin{tabular}{@{}l cc cc cc@{}}
\toprule
\textbf{Benchmark}
&                                         \makecell[c]{Qwen3-4B}        & \makecell[c]{Racka-4B} & \makecell[c]{Llama 3.1 8B} & \makecell[c]{ChatPULI} & \makecell[c]{Qwen2.5 7B} & \makecell[c]{PULI \\ Trio-Q Chat} \\
\midrule
HuMatchingFIB {\scriptsize (B acc.)}        & \textbf{48.83\tiny$\pm$0.90} & 35.54\tiny$\pm$1.08 & 30.49\tiny$\pm$0.98 & 28.20\tiny$\pm$0.67 & 31.52\tiny$\pm$0.38 & 42.60\tiny$\pm$0.58 \\
HuMatchingFIB {\scriptsize (Q acc.)}     & \textbf{6.84\tiny$\pm$1.16} & 3.10\tiny$\pm$0.32 & 3.96\tiny$\pm$0.57 & 3.67\tiny$\pm$0.30 & 1.15\tiny$\pm$0.30 & 3.89\tiny$\pm$0.39 \\
HuStandardFIB {\scriptsize (B acc.)}        & 15.66\tiny$\pm$1.50 & 15.02\tiny$\pm$1.44 & 13.81\tiny$\pm$1.85 & \textbf{24.48\tiny$\pm$0.83} & 7.15\tiny$\pm$1.01 & 18.38\tiny$\pm$0.62 \\
HuStandardFIB {\scriptsize (Q acc.)}     & \textbf{4.30\tiny$\pm$0.76} & 3.23\tiny$\pm$1.32 & 1.72\tiny$\pm$0.59 & 1.51\tiny$\pm$0.59 & 0.65\tiny$\pm$0.59 & 0.86\tiny$\pm$1.40 \\
HuProverbRea {\scriptsize Acc. (2CQ)}               & 80.48\tiny$\pm$0.52 & 76.63\tiny$\pm$0.84 & 72.41\tiny$\pm$0.64 & 78.54\tiny$\pm$0.43 & 70.73\tiny$\pm$0.16 & \textbf{83.86\tiny$\pm$0.27} \\
HuProverbRea {\scriptsize Acc. (OE)}        & 61.50 & 58.77 & 63.00 & 74.80 & 48.90 & \textbf{76.21} \\
HuSimpleQA {\scriptsize Acc.}     & 8.43 & 8.66 & 11.14 & \textbf{17.40} & 4.56 & 11.11 \\
HuWildBench {\scriptsize (WBScore)}       & 59.90 & 13.89 & 61.33 & 65.91 & 47.67 & \textbf{68.97} \\
\bottomrule
\end{tabular}
\caption{Results of the base models and their Hungarian-adapted variants on OpenHuEval. Reasoning models are run with thinking enabled. Values are means $\pm$ standard deviations across five runs, except for the LLM-judged metrics (HuSimpleQA, HuProverbRea (OE) and HuWildBench), which were evaluated in a single run. The highest score in each row is shown in bold.}
\label{tab:results_openhueval}
\end{table*}

The \textsc{OpenHuEval} results in Table~\ref{tab:results_openhueval} further confirm the gap between \textsc{Qwen3-4B} and \textsc{Racka-4B}.
The \textsc{Qwen3-4B} model ranks first on three of the eight metrics, while \textsc{Racka-4B} does not meaningfully outperform its base model on any metric. Its only higher score is on \textsc{HuSimpleQA} (8.66 vs.\ 8.43), which is negligible, and on both \textsc{HuStandardFIB} metrics the difference falls within the standard deviation.
Averaged across the accuracy-based metrics (i.e., all except \textsc{HuWildBench}), it is behind by 3.6 percentage points. Among these, the largest drop is on \textsc{HuMatchingFIB}~(B) (13.3 points).
The drop on \textsc{HuWildBench} is even larger (from 59.90 to 13.89).
\textsc{ChatPULI} and \textsc{PULI Trio-Q Chat} achieve the best scores on two and three metrics, respectively.
\textsc{PULI Trio-Q Chat} improves upon its base model on seven of the eight metrics, and on the remaining one (\textsc{HuStandardFIB}~(Q)) the difference falls within the standard deviation.
For \textsc{ChatPULI}, the pattern is more mixed. Among the metrics evaluated over five runs, it outperforms \textsc{Llama 3.1 8B} on two, falls behind on one, and the remaining two differences fall within the standard deviation. On all three single-run metrics, however, it scores higher, by 4.6 to 11.8 points.
\begin{table}[ht]
\centering
\setlength{\tabcolsep}{4pt}
\begin{tabular}{@{}l lll ll@{}}
\toprule
& \multicolumn{3}{c}{  \textbf{Qwen3-4B}}                                                                       & \multicolumn{2}{c}{\textbf{Racka-4B}} \\
\cmidrule(lr){2-4}\cmidrule(lr){5-6}
\textbf{Task}          & \textit{base}                          & \textit{w/o thinking}         & \textit{thinking}             & \textit{w/o thinking}             & \textit{thinking} \\
\midrule
MMLU                   & 16.49\tiny$\pm$0.24                    & 37.08\tiny$\pm$0.22           & \textbf{70.67 \tiny$\pm$0.18}$^*$ & 41.99\tiny$\pm$0.22               & 34.10\tiny$\pm$0.32 \\
TruthfulQA             & 28.52\tiny$\pm$1.70                    & 30.43\tiny$\pm$1.90           & \textbf{62.43\tiny$\pm$1.79}$^*$  & 16.47\tiny$\pm$1.26               & 26.85\tiny$\pm$1.36 \\
Spelling               & 87.22\tiny$\pm$9.30                    & 94.78\tiny$\pm$0.20           & 94.08\tiny$\pm$0.22           & 97.60\tiny$\pm$0.22               & \textbf{97.83\tiny$\pm$1.11}$^*$ \\
Prompt alignment       & 19.20\tiny$\pm$3.83                    & 70.00\tiny$\pm$1.41           & \textbf{74.00\tiny$\pm$2.74}$^*$  & 39.20\tiny$\pm$4.09               & 32.00\tiny$\pm$3.08 \\
Readability            & 39.86\tiny$\pm$10.69                   & \textbf{73.96\tiny$\pm$4.99}  & 68.88\tiny$\pm$2.61           & 61.56\tiny$\pm$2.80               & 33.68\tiny$\pm$8.06 \\
Bias                   & 89.80\tiny$\pm$2.04                    & 76.12\tiny$\pm$3.35           & \textbf{90.21\tiny$\pm$2.94}  & 62.04\tiny$\pm$2.44               & 24.49\tiny$\pm$5.72 \\
Toxicity               & 90.40\tiny$\pm$2.70                    & \textbf{90.80\tiny$\pm$1.48}  & 90.20\tiny$\pm$2.17           & 60.60\tiny$\pm$6.77               & 9.00\tiny$\pm$1.58  \\
Faithfulness           & 95.20\tiny$\pm$2.77                    & \textbf{99.80\tiny$\pm$0.45}$^*$ & \textbf{99.80\tiny$\pm$0.45}$^*$  & 73.00\tiny$\pm$4.00               & 41.60\tiny$\pm$6.54 \\
Summarization          & 20.00\tiny$\pm$3.65                    & \textbf{88.57\tiny$\pm$3.71}$^*$ & 84.90\tiny$\pm$3.10           & 82.04\tiny$\pm$5.48               & 21.63\tiny$\pm$3.98 \\
Answer relevancy       & 13.20\tiny$\pm$1.64                    & 47.00\tiny$\pm$3.54           & \textbf{66.00\tiny$\pm$5.29}   & 35.80\tiny$\pm$4.15       & 28.60\tiny$\pm$5.50 \\
Cola                   & 57.99\tiny$\pm$19.42                   & 90.59\tiny$\pm$1.76           & 85.63\tiny$\pm$1.35           & 97.60\tiny$\pm$2.46               & \textbf{98.41\tiny$\pm$1.13} \\
\bottomrule
\end{tabular}
\caption{HuGME results for the \textsc{Qwen3-4B} and \textsc{Racka-4B} models, with and without reasoning enabled. Mean $\pm$ standard deviation over $5$ runs. Best average result per row within this table in bold. Asterisks (*) mark the best result across Tables~\ref{tab:hugme_racka} and~\ref{tab:hugme_puli}.}
\label{tab:hugme_racka}
\end{table}

Results on \textsc{HuGME} show that reasoning mode helped \textsc{Qwen3-4B} -- as expected -- on knowledge- and reasoning intensive tasks, and \textsc{Racka-4B} scored lower on nine of the eleven tasks.
Comparing the two models without reasoning, a similar tendency can be observed.
\textsc{Racka-4B} performed better on only three tasks.
The pattern of improvements and degradations is consistent with the training objective.
Continued pretraining on raw text optimizes for next-token prediction, an objective that directly improves spelling and grammaticality.
Prompt alignment, faithfulness and safety behaviour like bias and toxicity, by contrast, were acquired during the post-training of the base model, and were not reinforced afterwards.

A similar pattern is reported by \citet{khade}, who applied LoRA fine-tuning to Marathi language.
One of their main findings is that target-language generation improved while reasoning degraded.
The authors also point out that although automatic scores dropped, human evaluators often favoured the adapted model. They conclude that evaluation methodology needs improvement, a caution that applies to the interpretation of our own results as well.
Three of the \textsc{HuGME} tasks -- toxicity, bias and faithfulness -- reward empty, repetitive and off-topic responses with high scores\footnote{We filtered out empty and repetitive outputs in this analysis, but off-topic responses still remained.}.
This is clearest in the \textsc{PULI-Trio-Q Base} column, which scores $0.00$ on summarization and $1.20$ on answer relevance while obtaining the table's best toxicity and faithfulness scores and a high bias score.
These metrics do not penalize this type of generation failure.

Despite these results, the use of LoRA is justified on several grounds.
In the case of \textsc{Racka}, only $12.5\%$ of the total parameters were trained, which made the adaptation of a 4B model feasible.
The argument that LoRA mitigates forgetting also has support in the literature \citep{biderman2024loralearnsforgets}.
The underlying cause therefore cannot be determined from the available data.
The limitations of LoRA, the composition of the training data and the absence of post-training may all have contributed, and separating these factors would require ablation experiments.

\begin{table}[ht]
\centering
\setlength{\tabcolsep}{4pt}
\begin{tabular}{@{}l lll lll@{}}
\toprule
\textbf{Task}    & Llama-3.1                     & PULI Base                             & ChatPULI                    & Qwen2.5             & PULI-Trio-Q Base     & PULI-Trio-Q Chat \\
\midrule
MMLU             & 34.19\tiny$\pm$0.13           & 19.84\tiny$\pm$0.30                   & 43.46\tiny$\pm$0.72          & 48.52\tiny$\pm$0.15 & 4.67\tiny$\pm$0.16           & \textbf{59.55\tiny$\pm$0.21} \\
TruthfulQA       & 20.59\tiny$\pm$2.96           & \textbf{54.42\tiny$\pm$5.37}          & 29.25\tiny$\pm$1.21          & 37.93\tiny$\pm$0.85 & 21.65\tiny$\pm$1.19          & 46.55\tiny$\pm$0.92 \\
Spelling         & 94.79\tiny$\pm$0.90           & \textbf{96.65\tiny$\pm$0.68}          & 95.21\tiny$\pm$0.41          & 92.26\tiny$\pm$1.24 & 95.59\tiny$\pm$0.79          & 94.70\tiny$\pm$0.25 \\
Prompt alignment & 55.80\tiny$\pm$7.19           & 38.20\tiny$\pm$2.59                   & 64.00\tiny$\pm$4.69          & 56.20\tiny$\pm$6.06 & 2.60\tiny$\pm$1.34           & \textbf{69.40\tiny$\pm$2.19} \\
Readability      & 71.72\tiny$\pm$1.49           & 71.62\tiny$\pm$6.95                   & \textbf{75.80\tiny$\pm$2.85} & 74.88\tiny$\pm$2.57 & 36.42\tiny$\pm$8.68          & 74.56\tiny$\pm$2.11 \\
Bias             & 89.18\tiny$\pm$3.43           & \textbf{93.27\tiny$\pm$2.76}$^*$      & 92.45\tiny$\pm$2.76          & 88.57\tiny$\pm$3.64 & 89.97\tiny$\pm$19.19         & 85.10\tiny$\pm$6.95 \\
Toxicity         & 92.20\tiny$\pm$4.60           & 94.40\tiny$\pm$2.07                   & 93.80\tiny$\pm$2.17          & 96.00\tiny$\pm$1.58 & \textbf{98.80\tiny$\pm$0.84}$^*$ & 90.20\tiny$\pm$3.56 \\
Faithfulness     & 96.40\tiny$\pm$1.82           & 96.60\tiny$\pm$2.30                   & 98.60\tiny$\pm$1.14          & 98.20\tiny$\pm$1.30 & \textbf{99.50\tiny$\pm$0.55}         & 99.40\tiny$\pm$0.89 \\
Summarization    & \textbf{75.92\tiny$\pm$13.18} & 56.73\tiny$\pm$4.87                   & 63.26\tiny$\pm$4.09          & 64.90\tiny$\pm$12.8 & 0.00\tiny$\pm$0.00           & 40.41\tiny$\pm$4.19 \\
Answer relevancy & 65.40\tiny$\pm$11.84          & 44.00\tiny$\pm$3.46                   & \textbf{80.00\tiny$\pm$4.74}$^*$ & 27.20\tiny$\pm$7.73 & 1.20\tiny$\pm$0.84           & 76.00\tiny$\pm$6.04 \\
Cola             & 93.23\tiny$\pm$1.39           & 93.74\tiny$\pm$2.12                   & \textbf{99.29\tiny$\pm$0.35}$^*$ & 43.73\tiny$\pm$4.10 & 85.74\tiny$\pm$4.38          & 97.08\tiny$\pm$1.91 \\
\bottomrule
\end{tabular}
\caption{\textsc{HuGME} results for the PULI models and the base models from which they were adapted. Model abbreviations: Llama-3.1 (Llama-3.1-8B-Instruct), PULI Base (PULI-LlumiX-Llama-3.1), ChatPULI (PULI-LlumiX-Llama-3.1 Chat) and Qwen2.5 (Qwen2.5-7B-Instruct). Mean $\pm$ standard deviation over $5$ runs. Best average result per row within this table in bold. Asterisks (*) mark the best result across Tables~\ref{tab:hugme_racka} and~\ref{tab:hugme_puli}.}
\label{tab:hugme_puli}
\end{table}




The results for the \textsc{PULI} models can be divided into two stages.
The first is the state after continued pretraining but before supervised fine-tuning (SFT), where the two branches behave differently.
On the \textsc{Qwen2.5} branch, the performance of the intermediate checkpoint drops sharply.
\textsc{PULI Trio-Q Base} scores $4.67$ on HuMMLU against the $48.52$ of the base model, $2.60$ on the prompt alignment task compared with
$56.20$, and $0.00$ on summarization.
On the \textsc{Llama 3.1} branch, the metrics give a mixed picture.
\textsc{PULI Base} declines on HuMMLU, prompt alignment and answer relevance, but outperforms the base model on TruthfulQA, spelling and the safety metrics.
The second stage, after SFT, the models appears to recover the lost performance and, in several cases, surpass the original models.
For instance, \textsc{PULI-Trio-Q Chat} scores $59.55$ on HuMMLU and $69.40$ on prompt alignment, and \textsc{ChatPULI} reaches
$80.00$ on answer relevancy.

For the \textsc{PULI} models, the adaptation is largely positive according to the \textsc{HuGME} benchmark, but the intermediate stage comes at a cost in capability that fine-tuning later recovers.
This raises the question of how continued pretraining contributed to the final outcome.
A control experiment would be informative, in which the model is trained with supervised fine-tuning alone, without continued pretraining.

\section{Gaps in the training data pipeline}
\label{sec:training}

The importance of high-quality training data is correctly emphasised in all three publications, consistent with findings from large-scale curation efforts showing that these decisions have a greater effect on downstream performance than additional raw volume \citep{penedo2025fineweb2}. Furthermore, recent practice treats the compute-optimal ratio \citep{hoffmann2022} as a lower bound rather than a target, particularly for smaller models, where extended training could lead to increased accuracy \citep{devries2023chinchilla_analysis}.

All three studies derive their Hungarian web corpora from the same processing pipeline, introduced by Dávid Nemeskey \citep{nemeskey2020}. At high-level, the pipeline consists of downloading webpages from selected top-level domains (from Common-Crawl\footnote{\url{https://commoncrawl.org/}} and/or other corpuses), boilerplate removal with JusText library\footnote{\url{https://github.com/miso-belica/justext}} supplemented by hand-crafted regular expressions, document-level language identification, length-based filtering, with URL-, document- and paragraph-level deduplication. This remains a sound foundation, and \citet{otp_2025} extend it by adding additional top-level domains to capture Hungarian content outside the .hu domain.

However, the pipeline reflects the state of the art of 2020, and several stages that have since become standard in corpus construction are absent from all three studies. These include model-based quality filtering, where a classifier is trained on annotated samples and scores documents for educational or informational value \citep{penedo2024finewebdatasets,allal2025smollm2}; domain and topic classification, which allows the composition of the corpus to be characterised and controlled; and benchmark decontamination, that is, the removal of documents overlapping with evaluation sets, a procedure already established at the time by \citep{brown2020languagemodelsfewshotlearners}; and content-level filtering for personally identifiable information or toxicity material \citep{soldaini2024dolma}.

More importantly, none of the publications reports ablation studies linking their corpus construction choices to downstream performance, as, for instance, in the development of \textsc{SmolLM2} \citep{allal2025smollm2}. Absent such evidence, the curation decisions remain, in the words of one of the papers, ``a strategic design choice aimed at achieving robust language adaptation'' \citep[Sec.~3.1]{racka_2026},  rather than a validated one.

Another concern is how the corpora are described. \textsc{Racka} and \textsc{OTP} both document the origin of their Hungarian material in detail, reporting document and token counts per source \citep[Appendix~B]{racka_2026}, \citep[Table~II]{otp_2025}. ChatPULI reports only aggregate figures for its Hungarian component \citep[Table~1]{chatpuli_2025}. However, origin is not composition. None of the three publications reports the distribution of domains or subject areas within its corpus, nor the proportion of scientific, technical or mathematical content. These categories have become explicit design variables elsewhere, to the point that dedicated resources are constructed where the available material proves insufficient \citep{allal2025smollm2}. Repository-level labels such as EPA\footnote{\url{https://epa.oszk.hu/}} or MEK\footnote{\url{https://mek.oszk.hu/hu/}} identify where a document was obtained, not what it contains. These collections span peer-reviewed research, classical literature, textbooks and digitised material, while a single label subsumes all of it.

Hungarian pretraining data does exist in usable quantities, but the post-training stage is basically absent. To our knowledge, the only publicly documented Hungarian chat and instruction-tuning dataset is the one assembled at NYTK, comprising $44,626$ segments \citep{chatpuli_2025}, of which $7,344$ derive from the HuLU benchmark (a clear example of contamination). No Hungarian preference-tuning dataset has been released, and consequently no model (developed by Hungarian research teams) has undergone preference-tuning either with RLHF or RLVR. \textsc{Racka} is published as a base model, with supervised fine-tuning and preference tuning explicitly deferred to future work \citep{racka_2026}. Four years after RLHF became standard practice \citep{ouyang2022} and more than a year after RLVR recipes were released openly \citep{deepseek_r1_2025, lambert2025tulu3pushingfrontiers}, the post-training stage of the pipeline remains effectively unaddressed for Hungarian. A part of the explanation for this is legitimate. High-quality instruction and preference data is expensive to collect, human annotation requires funding, and much of the text that would serve as source material is under copyright.

Although in recent years synthetic data generation has become a mature and viable alternative with the right base model \citep{wang2023selfinstruct, xu2025wizardlmempoweringlargepretrained, abdin2024phi4technicalreport, benallal2024cosmopedia, nvidia2024nemotron4}, a synthetic corpus can be no better than the model that produced it. For Hungarian, two conditions must meet at once: the proficiency in the language, and its licence must allow its outputs to be used for training.
Selecting among these candidates is the easier part of the problem, verifying is the challenging part. Hungarian currently has no dedicated resource for measuring the properties that determine whether generated data is worth training on, for instance naturalness as opposed to translationese, and diversity across generations. The existing \textsc{HuGME} benchmark \citep{ligetinagy2025hugme} is a starting point, and some of its tasks are useful, but it does not fully substitute for this. The obvious, but expensive solution would be to measure the downstream performance of models trained on synthetic data generated by each candidate \citep{penedo2024finewebdatasets}.

\section{Missing ablations and data mixture in model training}
\label{sec:ablations}

Beyond data composition, the description of the training phases and of ablation experiments is also missing, or rather only a basic form of these can be observed.
\citet{otp_2025} trained their models in two phases.
The first phase used $500$B tokens with a $2$k context window, the second a further $140$B tokens with an $8$k window.
The Hungarian subset was oversampled four times, an equal proportion of English data was added, and the resulting datasets were alternated during training.
\citet{chatpuli_2025} concluded continued pretraining with a Hungarian-only stage, introduced because of an ``increased number of grammatical errors'', though no measurements are reported to support this.
\citet{racka_2026} also oversampled certain parts of the corpus, but it is unclear whether the different sources are distributed evenly across training or appear in consecutive blocks.

Additionaly, none of the three publications reports ablation studies to justify their data mixture choices to downstream performance, like in \cite{raffel2023exploringlimitstransferlearning, hu2024minicp} and \cite{bakouch2025smollm3}. \textsc{ChatPULI}'s final stage is the closest analogue to a cooldown, and its effect is not quantified.

\section{Measuring forgetting and capability loss}
\label{sec:forgetting}

The \textsc{Racka} team use low-rank adaptation (LoRA) \citep{hu2021lora}, on the grounds that the method should ``inherently aid in overcoming catastrophic forgetting of the base model's knowledge'' \citep[Sec.~2.2]{racka_2026}.
However, the cited study reports a relative rather than an absolute effect \citep{li2024forgetting}.
LoRA retains more of the base model's capabilities than full-parameter continual pretraining does, but it still underperforms the starting model, \textsc{Llama-2-7B-chat} \citep{touvron2023llama2}, on most metrics.
The method therefore mitigates the damage rather than eliminating it.
In addition to PEFT, the \textsc{Racka} work follows the approach of \citet{csaki2023}, who mitigate catastrophic forgetting by extending the tokenizer vocabulary and tuning the data mixture.
That work adapted a pretrained model, whereas \textsc{Qwen3-4B} has also been tuned for instruction following, dialogue and reasoning, making its behaviour considerably more complex \citep{yang2025qwen3technicalreport}.
Continual pretraining is carried out on raw text, and the study does not test whether the behaviours established during post-training are preserved.
Their findings therefore do not imply that those capabilities survive adaptation.
More importantly, \citet{li2024forgetting} measured forgetting in the adapted model against the starting model.
The \textsc{Racka} authors appear to adopt (and extend) these conclusions without adopting the corresponding measurements.
No comparison against \textsc{Qwen3-4B}'s own evaluation suite (or a subset of it) is reported.

The same omission occurs in the case of \textsc{ChatPULI}, as the authors likewise do not measure the extent of capability loss.
They too start from a post-trained model, and they present the procedure as novel, stating that ``we found no previous research that applies continued pre-training on an instruct model for language adaptation''.
In fact, the study by \citet{li2024forgetting} does exactly this, adapting \textsc{Llama-2-7B-Chat} to Chinese.

\begin{figure*}[t]
\centering
\includegraphics[width=0.8\textwidth]{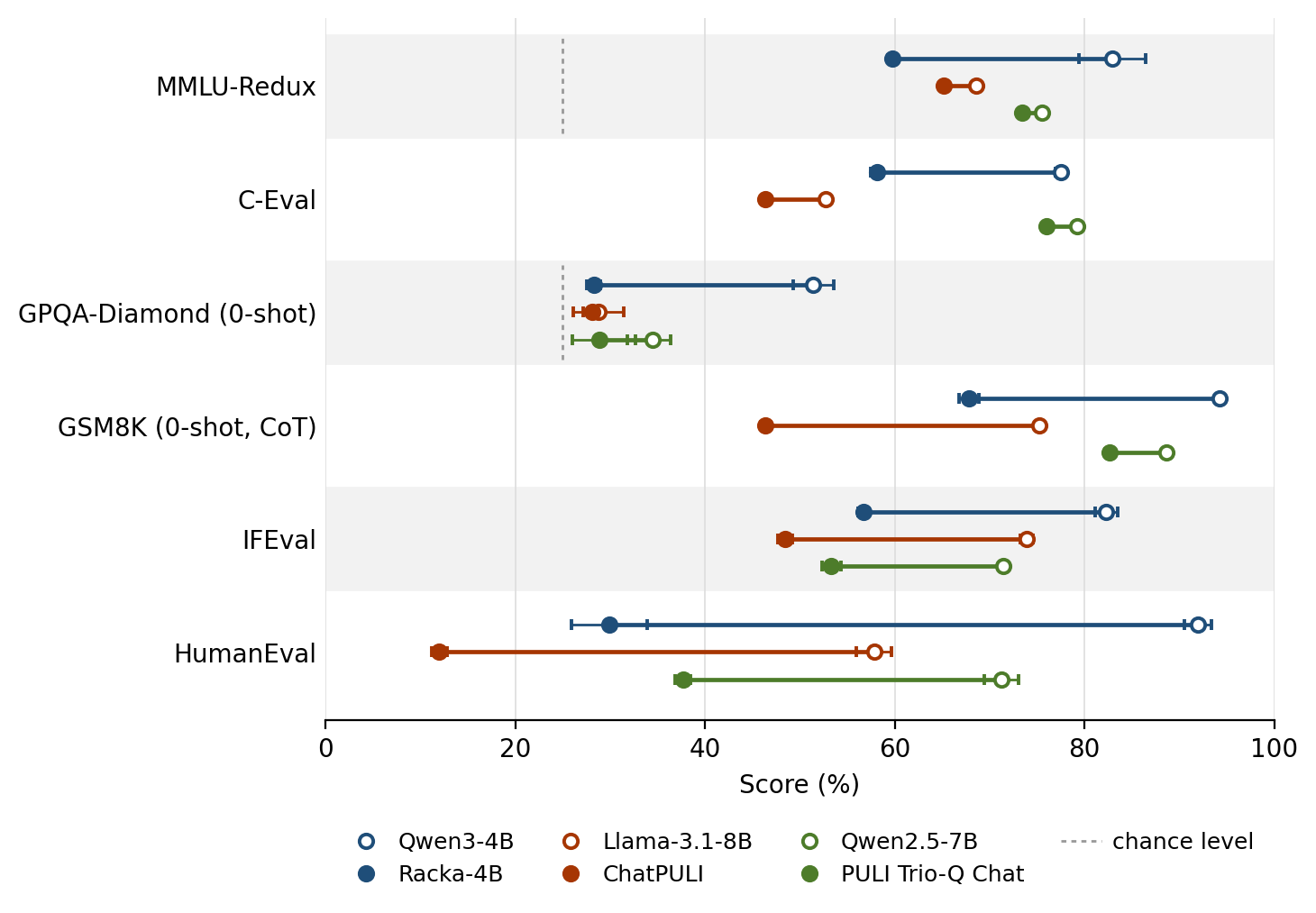}
\caption{Score differences between base and adapted models on the evaluated benchmarks. Open circles mark base models, filled circles adapted models. The dashed line marks chance level on the multiple-choice benchmarks.}
\label{fig:forgetting}
\end{figure*}

To measure forgetting, each adapted model was evaluated alongside its starting model on a subset of the base-model benchmarks.
The subset covers general knowledge (\textsc{MMLU-Redux} \citep{gema-etal-2025-done}, \textsc{C-Eval} \citep{huang2023ceval}), reasoning and mathematics (\textsc{GSM8K} \citep{cobbe2021gsm8k}, \textsc{GPQA-Diamond} \citep{rein2023gpqa}), instruction following (\textsc{IFEval} \citep{zhou2023ifeval}) and code generation (\textsc{HumanEval} \citep{chen2021humaneval}).
For the evaluation, we used version 0.4.13 of the \texttt{lm-evaluation-harness} library\footnote{\url{https://github.com/EleutherAI/lm-evaluation-harness/tree/main}}.
For \textsc{Qwen3-4B}, we applied the prompts recommended in the documentation\footnote{\url{https://huggingface.co/Qwen/Qwen3-4B}}, we used the same prompts for \textsc{Racka-4B}.
The results are shown in Table~\ref{tab:forgetting} and Figure~\ref{fig:forgetting}.

\begin{table*}[ht]
\centering
\setlength{\tabcolsep}{7pt}
\begin{tabular}{@{}l cc cc cc@{}}
\toprule
\textbf{Benchmark}
&                                           \makecell[c]{Qwen3-4B} & \makecell[c]{Racka-4B} & \makecell[c]{Llama 3.1 8B} & \makecell[c]{ChatPULI}   & \makecell[c]{Qwen2.5 7B}  & \makecell[c]{PULI \\ Trio-Q Chat} \\
\midrule
MMLU-Redux                                  & \textbf{82.91\tiny$\pm$3.49}  & 59.74\tiny$\pm$0.44    & 68.53\tiny$\pm$0.08        & 65.19\tiny$\pm$0.09      &  75.53\tiny$\pm$0.04      & 73.40\tiny$\pm$0.04 \\
C-Eval                                      & 77.49\tiny$\pm$0.64  & 58.10\tiny$\pm$0.62    & 52.72\tiny$\pm$0.18        & 46.32\tiny$\pm$0.19      & \textbf{79.24\tiny$\pm$0.04}       & 75.94\tiny$\pm$0.10 \\
\midrule
GPQA-Diamond {\scriptsize (0-shot)}         & \textbf{51.41\tiny$\pm$2.12} & 28.28\tiny$\pm$0.71 &  28.79\tiny$\pm$2.67             & 28.08\tiny$\pm$0.98 &  34.49\tiny$\pm$1.88 &  28.87\tiny$\pm$2.90 \\
GSM8K {\scriptsize (0-shot, CoT)}           & \textbf{94.27\tiny$\pm$0.29} & 67.82\tiny$\pm$1.07 & 75.25\tiny$\pm$0.28   & 46.37\tiny$\pm$0.49 &  88.58\tiny$\pm$0.18 &  82.59\tiny$\pm$0.43 \\
\midrule
IFEval {\scriptsize (strict prompt)}        & \textbf{82.29\tiny$\pm$1.21} &  56.69\tiny$\pm$0.58 &  73.90\tiny$\pm$0.66 &  48.43\tiny$\pm$0.77 &  71.42\tiny$\pm$0.21 &  53.31\tiny$\pm$1.03 \\
\midrule
HumanEval                                   & \textbf{91.95\tiny$\pm$1.46} & 29.88\tiny$\pm$3.98 &  57.80\tiny$\pm$1.85 & 11.95\tiny$\pm$0.8 &  71.22\tiny$\pm$1.80 & 37.68\tiny$\pm$0.80 \\
\bottomrule
\end{tabular}
\caption{Comparison among base models and their Hungarian adaptation on a subset of the benchmarks used for the base models. The reasoning models are set to thinking mode. Mean $\pm$ standard deviation over 5 runs. Best average result per row in bold.}
\label{tab:forgetting}
\end{table*}

Some form of capability loss is observable on every benchmark after continual pretraining, with the exception of \textsc{GPQA-Diamond}, where \textsc{ChatPULI} shows no substantial difference from \textsc{Llama 3.1-8B-Instruct}.
\textsc{Qwen3-4B} consistently surpasses \textsc{Racka-4B} on all examined benchmarks, with the largest gap on \textsc{HumanEval}.
\textsc{ChatPULI} and \textsc{PULI Trio-Q Chat} show a similar but smaller decline relative to their respective base models, the only exception is \textsc{ChatPULI} on \textsc{GSM8K}, where its decline is greater than that of \textsc{Racka-4B}.

Overall, all three adaptations show a trade-off between language adaptation and capability retention, though its extent varies by model and task type.

\section{Training cost and carbon footprint}
\label{sec:trainingcost}

The remarkable performance of recent language models has come at a substantial cost: high energy use, training cost and carbon emissions \citep{lacoste2019carbonemissions}, which developers increasingly disclose alongside their results \citep{touvron2023llamaopenefficientfoundation, bellagente2024stablelm216b}.

However, reporting among the three Hungarian research is uneven. \textsc{Racka} is the only one to provide an explicit estimate: $1,290$ kg CO\textsubscript{2}eq for a run of $287$ hours on $64$ NVIDIA A100 (40 GB) GPUs, and attributing the comparatively low figure to the design of the host facility. Unfortunately, they did not provide the calculation methodology behind this figure, or specify what aspects of the facility's design are responsible for the claimed efficiency. \textsc{ChatPULI} only reports the hardware (4 NVIDIA A100 GPUs with 80 GB VRAM) and training configuration\footnote{The study uses \texttt{float32} precision for the \textsc{Qwen2.5}-based model, but does not specify whether TF32 was enabled. The distinction is important, as on A100 hardware, the \texttt{float32} attains $19.5$ TFLOPS, whereas TF32 reaches $156$ TFLOPS. Under \texttt{float32}, the estimated wall-clock time exceeds $17,000$ hours, which is not consistent with the scope and timeline of the project. Therefore, we assume TF32 in our estimate.}. The \textsc{OTP} publication states that training was performed on $96$ SambaNova SN10 RDUs and that the $13$B model training took four months to complete, but neither the paper nor the source it cites \citep{peng2024} reports power consumption or energy use.

It would be useful to have a more complete picture of the environmental and financial costs of training these models, especially given the growing concern about the sustainability of AI research and inference. Therefore, in line with \citet{lacoste2019carbonemissions} and \citet{wu2022sustainablea}, we attempt to estimate the energy consumption, carbon emissions and training cost of the Hungarian models based on the information provided in their publications. Energy consumption are computed as:
\begin{equation}
    \text{Wh}=\text{GPU-hours} \times \text{GPU power consumption } \times \text{PUE}.
\end{equation}

We initially considered a Power Usage Effectiveness (PUE) of 1.6, the reported average for European data centres \citep{jrc2023eucode}, but Komondor has been reported to rank favourably on the Green500 list of energy-efficient supercomputers \citep{hpchu2025komondor}, suggesting a facility PUE substantially below the EU average. As we could not find an official PUE for Komondor, we adopt a value of $1.1$, the same as used in \citet{touvron2023llamaopenefficientfoundation}.

To ensure a consistent basis for comparison across models, we apply a fixed carbon intensity factor of $0.154$ kg CO\textsubscript{2}eq/kWh, per the European Environment Agency's official indicator \citep{eea2024ghgintensity}, regardless of the actual server location. This gives the following formula for calculating carbon emission:
\begin{equation}
    \mathrm{tCO}_2 \mathrm{eq}=\mathrm{MWh} \times 0.154 .
\end{equation}

For training cost, we adopt the official Komondor price list\footnote{\url{https://docs.hpc.dkf.hu/first-steps/pricing.html}}, which specifies a rate of $1.115$ EUR per GPU-hour for the GPU and AI partitions:
\begin{equation}
\text{Total Training Cost (EUR)} = \text{GPU-hours} \times 1.115.
\end{equation}

As the \textsc{Racka} and \textsc{PULI} model training runs\footnote{In practice, the \textsc{PULI} models were trained on NYTUD's own servers rather than on Komondor.} were conducted as part of a supported project, the value reported here represents a notional cost, i.e. the amount that would have been invoiced under Komondor's standard commercial pricing. For consistency, we apply the same formula and price to these two models, enabling a fair comparison\footnote{The reported training footprints typically cover only the final training run, excluding the cost of experimentation. The examined studies share this limitation.}. To the best of our knowledge, SambaNova has not publicly disclosed the power consumption of the SN10 RDU. Consequently, our estimations remains incomplete for the \textsc{OTP} models. The results of our calculations are summarized in Table~\ref{tab:carbon-emissions}.

\begin{table}[htbp]
\centering
\resizebox{\textwidth}{!}{%
\begin{tabular}{llccccc}
\toprule
 & Accelerator & \shortstack{Power\\consumption} & \shortstack{Accelerator \\ hours} & \shortstack{Total power\\consumption} & \shortstack{Carbon emitted\\(tCO$_2$eq)} & \shortstack{Training cost \\ (EUR)} \\
\midrule
Racka-4B            & A100 (40GB) GPU & 400 W & 18,386 & 8.08 MWh & 1.24 & 20,500.39 \\
\midrule
OTP-1.5B            & SN10 RDU        & -     & -        & - & - & -\\
OTP-13B             & SN10 RDU        & -     & 276,480  & - & - & - \\
\midrule
PULI Base           & A100 (80GB) GPU & 400 W & 3,875 & 1.72 MWh  & 0.26 & 4,321 \\
PULI Trio Q Base    & A100 (80GB) GPU & 400 W & 8,845 & 3.89 MWh & 0.60 & 9,862 \\
\bottomrule
\end{tabular}%
}
\caption{Estimated energy usage, carbon footprint and training cost by model. Only the continual pretraining and pretraining stages are considered.}
\label{tab:carbon-emissions}
\end{table}

Reconstructing \textsc{Racka}'s carbon emissions yields $1,244.6 \text{kg CO}_2\text{eq}$, within $3.5\%$ of their reported $1,290 \text{kg CO}_2\text{eq}$. This suggests their figure is consistent with an efficient hyperscale facility PUE rather than the EU data-center average of $1.6$ \citep{jrc2023eucode}, which would overshoot their reported total by $\sim 30\%$.
Similarly, the \textsc{ChatPULI} publication does not report training duration, energy consumption, emissions, or monetary cost. Consequently, our estimates for the \textsc{PULI} models are reconstructed from the reported training statistics, for which we use the calculations in \citet{hoffmann2022trainingcomputeoptimallargelanguage, chowdhery2022palm, kaplan2020scalinglawsneurallanguage}.

Publishing the above factors and detailing the computational methodology is critical not only from an environmental perspective but also -- similar to performance -- enables comparison across different models.

\section{Conclusion}

In recent years, language model development has become one of the fastest-growing research areas.
In Hungary, three major projects aimed to develop generative language models.
While these are important attempts at modelling the Hungarian language, our analysis reveals a number of methodological shortcomings that limit the interpretability of their results.
Moreover, all three have remained within the earlier part of the modern language modelling timeline presented in Appendix~\ref{appendix:stateofllms}.

A shared characteristic of the three projects is that each built up the infrastructure required for model development largely independently of the others.
This includes corpus construction as well as the creation and validation of benchmarks.
Since each of these tasks demands substantial resources on its own, and the Hungarian NLP community is relatively small, duplicating the effort in parallel increases the probability of methodological errors, which manifest, for example, in flawed evaluation.
Internationally, these responsibilities are distributed across the community and between research groups.

Although full openness may not be achievable, closer coordination between Hungarian research groups and institutions would already be a significant improvement on current practice.

\printbibliography

\appendix

\section{Appendix}

\subsection{The evolution of language models since the transformer}
\label{appendix:stateofllms}

One starting point for this process was the transformer architecture introduced by \citet{vaswani2023attentionneed}.
It was originally designed as an encoder-decoder model for machine translation. The encoder-only \textsc{BERT} appeared in 2019 \citep{devlin2019bertpretrainingdeepbidirectional}, followed by the autoregressive \textsc{GPT-3} \citep{brown2020languagemodelsfewshotlearners} in 2020.
These models established the pretraining and fine-tuning approach, together with the scaling of model and data size.
In early 2022, \textsc{InstructGPT} was released, tuned for instruction following with reinforcement learning from human feedback (RLHF) \citep{ouyang2022traininglanguagemodelsfollow}.
\textsc{ChatGPT} followed in November of the same year \citep{openai2022chatgpt}, making language models widely known outside the research community.
Other modalities were integrated in parallel.
\textsc{CLIP} \citep{radford2021} established a shared representation space for images and text, and \textsc{Flamingo} \citep{alayrac2022flamingovisuallanguagemodel} followed in 2022 as the first model to couple a pretrained language model with a vision encoder, enabling few-shot task solving on image-text input.
In the audio domain, \textsc{Whisper} \citep{radford2022robustspeechrecognitionlargescale} advanced speech recognition at the end of 2022, while \textsc{AudioLM} \citep{borsos2023audiolmlanguagemodelingapproach} carried the language modelling approach over to audio generation.
Tool use and interaction with the environment followed. \textsc{Toolformer} \citep{schick2023toolformerlanguagemodelsteach} and the \textsc{ReAct} framework \citep{yao2023reactsynergizingreasoningacting} were among the first steps in this direction, both appearing in 2023.
In September 2024, OpenAI released \textsc{o1}, generally regarded as the first widely available reasoning large language model \citep{openai2024o1preview}.
In January 2025, DeepSeek AI published \textsc{DeepSeek-R1} together with a detailed technical report showing how reasoning ability can be developed through reinforcement learning from verifiable rewards (RLVR) \citep{Guo_2025}.
A further shift began in 2024 and developed through 2025, moving the focus from chat to agents.\footnote{The term agent has two meanings. The classical, formal definition takes an agent to be an entity that perceives its environment through sensors and acts upon it through actuators \citep{russell2010aima}. In the context of large language models the term is narrower. A system in which the LLM dynamically directs its own processes and tool use, exercising control over how it pursues its tasks \citep{anthropic2024agents}.}
These systems are called agentic models, and the reasoning they perform is referred to as agentic reasoning. Models that plan, decide and act through interaction with an environment (e.g. a computer) \citep{anthropic2024computeruse, deepmind2025computeruse, openai2025operator}.
The change affected not only the models but the engineering layer around them.
A model's capabilities depend not only on the model itself, but on the environment and tooling built around it, i.e. the harness \citep{karten2026continualharnessonlineadaptation, lin2026agenticharnessengineeringobservabilitydriven}.
\citet{yang2024sweagentagentcomputerinterfacesenable} formalize this as the agent-computer interface (ACI) and show that its design produces substantial performance difference even with the same underlying model.
These interfaces are now a standard component of agent systems \citep{karten2026continualharnessonlineadaptation}.
Alongside these developments came a rapid succession of architectural, hardware and algorithmic optimizations\footnote{The works cited are not necessarily the earliest in each direction. The aim was to sketch a timeline of the major developments, without claim to completeness.}.

\subsection{Sampling parameters for HuGME benchmark}
\label{app:sampling_params_hugme}

Across all benchmarks and tasks, we set the maximum generation length to $8192$ tokens.
It turned out that reasoning models often exceeded this limit before producing a final answer.
We observed two failure modes in the outputs: repetitive text, and a missing closing \texttt{</think>} tag.
Both were treated as empty responses.
For example, on the MMLU task of \textsc{HuGME}, \textsc{Racka-4B} failed to close its reasoning $996.7$ times on average across $5$ runs, compared with $315.4$ times for \textsc{Qwen3-4B}.
The effect is even more pronounced on \textsc{HuWildBench}: $1{,}013$ of the $1{,}154$ responses from \textsc{Racka-4B} lacked the closing \texttt{</think>} tag.
On the remaining tasks, \textsc{Qwen3-4B} showed this behaviour in fewer than $1\%$ of cases, but it was persistent for \textsc{Racka-4B}.

The \textsc{Racka} paper notes that ``in many cases, the presence of reasoning traces resulted in a parsing error or the model running out of the allotted context'' \citep[Appendix~C]{racka_2026}.
However, the authors do not specify which models or tasks this observation refers to.
Our results suggest that it applies primarily to \textsc{Racka-4B}.
Unfortunately, extending the limit to $16$K or $32$K tokens was not feasible within our research budget.
The evaluations reported here cost $675.76$ USD in total.

Sampling parameters were configured as follows. For \textsc{Qwen3-4B} and \textsc{Racka-4B}, we adopted the values recommended in the Qwen3 documentation. In non-thinking mode, a temperature of $0.7$, top-p value of $0.8$. In thinking mode, we set the temperature to $0.6$, top-p to $0.95$. Additionally, top-k value of $20$ and min-p value of $0$ was used throughout. For all other models, we used temperature $0.8$, top-p $0.9$, and a repetition penalty of $1.1$.

\end{document}